\documentclass[10pt,twocolumn,letterpaper]{article}

\usepackage{cvpr}              
\usepackage{graphicx}
\usepackage{ifluatex}

\usepackage[accsupp]{axessibility}  
\usepackage{tikz}
\usepackage{comment}
\usepackage{amsmath,amssymb} 
\usepackage{color}
\usepackage{xcolor,colortbl}
\usepackage{framed,multirow}
\definecolor{LightCyan}{rgb}{0.65,.65,1}
\newcommand{\Blue}[1]{\color{blue}\textbf{#1}}
\newcommand{\Red}[1]{\textcolor{black}}
\usepackage{bbm}
\usepackage{dsfont}
\newcolumntype{C}[1]{>{\centering\let\newline\\\arraybackslash\hspace{0pt}}m{#1}}
\definecolor{Gray}{gray}{0.85}
\usepackage[hyperfootnotes=false]{hyperref}
\hypersetup{
  colorlinks,
  citecolor=green,
  linkcolor=red,
  urlcolor=blue}
\hypersetup{
  citebordercolor=green,
  filebordercolor=blue,
  linkbordercolor=blue
}
\usepackage{subcaption}
\usepackage{amsmath}
\usepackage{amssymb}
\usepackage{booktabs}
\usepackage[capitalize]{cleveref}

\usepackage{hyperref}
\crefname{section}{Sec.}{Secs.}
\Crefname{section}{Section}{Sections}
\Crefname{table}{Table}{Tables}
\crefname{table}{Tab.}{Tabs.}

\def\cvprPaperID{7} 
\def\confName{CVPR}
\def\confYear{2023}

\begin{document}
\newcommand{\name}{SCONE-GAN}

\title{\name: \underline{S}emantic \underline{Con}trastiv\underline{e} learning-based \underline{G}enerative \underline{A}dversarial \underline{N}etwork for an end-to-end image translation} 

\author{Iman Abbasnejad$^{1}$, Fabio Zambetta$^{1}$, Flora Salim$^{1,4}$, Timothy Wiley$^{1}$, \and Jeffrey Chan$^{1}$, Russell Gallagher$^{3}$, Ehsan Abbasnejad$^{2}$\\
$^{1}$RMIT University, $^{2}$Australian Institute for Machine Learning, $^{3}$ Rheinmetall Defence Australia, \\
$^{4}$UNSW Sydney University\\
\thanks{Iman Abbasnejad worked on this paper when he was working at RMIT University and Russell Gallagher worked on this project when he was at Rheinmetall Defence Australia.}
{\tt\small $^{1}$\{iman.abbasnejad,fabio.zambetta,timothy.wiley,jeffrey.chan\}@rmit.edu.au}\\
{\tt\small $^{2}$ehsan.abbasnejad@adelaide.edu.au,$^{3}$russell.gallagher@gmail.com,$^{4}$flora.salim@unsw.edu.au}\\
}
\maketitle

\begin{abstract}
\name~presents an end-to-end image translation, which is shown to be effective for learning to generate realistic and diverse scenery images. Most current image-to-image translation approaches are devised as two mappings: a translation from the source to target domain and another to represent its inverse. While successful in many applications, these approaches may suffer from generating trivial solutions with limited diversity. That is because these methods learn more frequent associations rather than the scene structures. To mitigate the problem, we propose~\name~that utilises graph convolutional networks to learn the objects dependencies, maintain the image structure and preserve its semantics while transferring images into the target domain. For more realistic and diverse image generation we introduce style reference image. We enforce the model to maximize the mutual information between the style image and output. The proposed method explicitly maximizes the mutual information between the related patches, thus encouraging the generator to produce more diverse images. We validate the proposed algorithm for image-to-image translation and stylizing outdoor images. Both qualitative and quantitative results demonstrate the effectiveness of our approach on four dataset.
\end{abstract}

\section{Introduction}

Generative Adversarial Networks (GANs)~\cite{goodfellow2014generative} are successful in generating high quality image samples from a random noise vector~\cite{karras2019style,karras2017progressive}. However, generating scenery images with high-fidelity in complex domains with multiple factors of variation using a noise vector remains challenging. One such application is generating an outdoor scene under different weather conditions in either conditional or unconditional case. For example, image translation (\ie conditional sampling) of an outdoor scene taken in summer (\ie domain $A$) into a realistic image of the same scene in winter (\ie domain $B$)~\cite{park2020contrastive,zhu2017unpaired}. \textcolor{black}{This image generation is useful for practical applications where it is necessarily to visualise scenery in different weather conditions, but it is not feasible or costly efficient to revisit and recapture the same location under multiple conditions.}
The majority of current state-of-the-art image-to-image translations try to compute a mapping function $f_{AB}$ and an \emph{estimate} of its inverse function $f_{BA}^{-1}$ which is combined with a cycle-consistency loss for synthesising image in the target domain~\cite{zhu2017unpaired,hoffman2018cycada,huang2018multimodal,mao2019mode}. 
\begin{figure}[t!]
\vspace{-25pt}
\centering
   \includegraphics[width=7.8cm, height= 5.0cm]{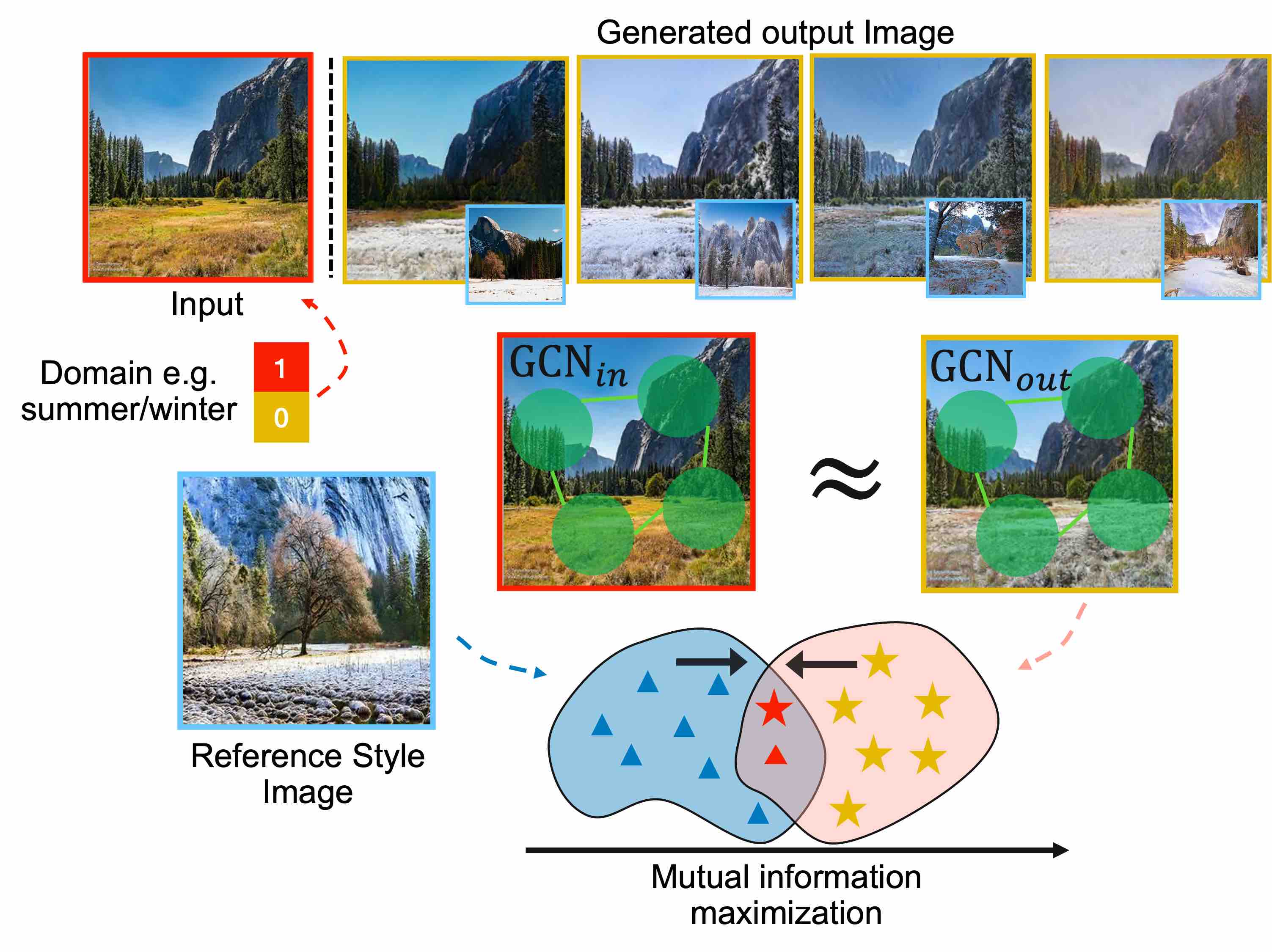}
   \vspace{-10pt}
   \caption{Overview of our proposed model. Given an input image (red rectangle), a reference style image (blue rectangle) and a domain (e.g. winter), our model can realistically generate images in the new domain (yellow rectangle). We establish a graph on the input and generated image for maintaining dependencies between objects (GCN$_{in}$ and GCN$_{out}$). We also maximize the mutual information between the reference and generated images for enhancing the diversity of generated images.}
\label{fig:model_pipeline}
\end{figure}

Despite the success of cycle-consistent loss, they have a major drawback. The reconstruction loss forces the generator to hide the information necessary to accurately reconstruct the input image due to a bias towards more frequent patterns rather than the structural consistencies~\cite{chu2017cyclegan}. The problem is particularly severe in high-frequency signals, such as outdoor scene synthesis, where the model must reconstruct too many features (including sky, clouds, mountains, etc.) and are unable to blend these elements together into a realistic image under the new domain. Therefore, cycle-consistent GANs can not be easily used to extract these information since all attributes such as objects, textures and colors are entangled. The lack of such control limits usage of image-to-image translation methods in many areas and may result in generating low quality with limited diversity images. 

Various studies have been conducted to address these limitations.~\cite{hoffman2018cycada} used global structural consistency through pixel cycle-consistency and semantic losses to adapt representations at both the pixel-level and feature-level for scenery image translation.~\cite{huang2018multimodal} decomposed the image representation into a content code, the information that should be preserved during translation, and a style code, the remaining variations that are not in the input image and should be mapped during translation, and generated images by combining content and style codes.~\cite{mao2019mode} demonstrated by maximizing the distance between generated images with respect to their corresponding latent codes, generator produces more diverse images.~\cite{lee2020drit++} decomposed the input image into a domain-invariant content space and a domain specific attribute space for generating diverse images.~\cite{park2020contrastive} maximized the mutual information between input and output patches using contrastive learning for learning the commonalities between two domains.  

Although some improvements have been made, the majority of previous techniques do not perform well on the complex outdoor scenery images. One common reason stems from the fact that, previous models did not fully represent an embedding space where the input image content is well preserved and fully mapped with the target domain information. To overcome these limitations, we propose a generative model that obtained state-of-the-art results on synthesizing outdoor scenery images given a domain and a style image. \textcolor{black}{Unlike previous models that manipulate images in the target domain using a noise variable, our model generates more realistic images where a user can control the output style image using a reference image.} Our model learns the semantic relationship between the objects in an end-to-end framework. To maintain the dependency between the objects we construct a graph convolutional network on the source and target images. For more realistic and diverse image creation, we encourage the network to maximize the mutual information between the reference and output images. We put weights on the mutual objects in the style image and output and penalize those that are different. We learn this through contrastive learning framework which has driven recent advances in learning representations~\cite{park2020contrastive,he2020momentum,chen2020simple}. Figure~\ref{fig:model_pipeline} illustrates an overview of our proposed model.
In this paper we make the following contributions:

\begin{itemize}
    \item We introduce \name~which can synthesize an image in the highly complex natural scenery in an end-to-end framework \textcolor{black}{using unpaired images}.
    
    \item Maintaining the content and image structure between the input and output images is critical in a realistic image translation. We utilise graph convolutional networks to build the object dependencies to preserve and maintain the image structure during translation.
    
    \item We demonstrate the efficiency of contrastive learning for a diverse image-to-image translation. We leverage the power of contrastive learning by putting more weight on the objects that are perceptually similar and penalize those objects that are different for a realistic and diverse image-to-image translation.
    
    \item \textcolor{black}{Controlling the style and content of generated images cannot be simply achieved by manipulating a noise vector. Therefore, we introduce a style-reference image that is used for stylizing the target image.}
\end{itemize}

\section{Related work}

The image-to-image translation approaches can be classified into two categories of paired~\cite{isola2017image,park2019semantic,mirza2014conditional,reed2016generative} and unpaired training methods~\cite{liu2016coupled,hoffman2018cycada,huang2018multimodal,lee2020drit++,abbasnejad2017affine,park2020contrastive}.
~\cite{isola2017image} introduced a General-purposed conditional GAN model~\cite{mirza2014conditional} to learn a one-side mapping function from the input images to target images.~\cite{odena2017conditional} synthesised images using random noise and a corresponding class label.~\cite{park2019semantic} learned to synthesise image given a segmentation mask and a style reference image. 
One common problem with the paired training approaches is that they only operate in a supervised setting when the paired training data is available.

Since providing a paired training sample is difficult, numerous methods have been introduced to tackle this limitation~\cite{aytar2017cross,liu2016coupled,hoffman2018cycada,huang2018multimodal,lee2020drit++,park2020contrastive,xie2020self}.~\cite{liu2016coupled} considered coupled-GANs for learning a joint probability distribution of two unpaired examples to learn translation.~\cite{aytar2017cross} used a weight-sharing strategy to learn a common representation across domains.~\cite{zhu2017unpaired} used a cycle-consistency loss~\cite{zhou2016learning} to learn a mapping from translation and reconstruction of the input and output images. 
~\cite{zhang2019harmonic} enforced a smoothness regularization term over the CycleGAN network to preserve consistent mappings during the translation. They showed for a better translation, the inherent property of samples should be preserved.~\cite{xie2020self} considered a self-supervised module to preserve image content during translation.

While maintaining image content is crucial for image translation, a successful translator should be able to accurately transform the image appearance conditioned on the target domain information. It has been shown that~\cite{hjelm2018learning,bachman2019learning,vondrick2018tracking,park2020contrastive} maximizing mutual information between the input data and image representations, can improve the visual representation learning and reconstruction performance.~\cite{vondrick2018tracking} showed for image colorization, maximizing the similarity between the reference and target images can improve the results.~\cite{park2020contrastive} maximized the mutual information between input and output patches for one-sided image translation, and obtained better results. 

In this work we build our unpaired image-to-image translator based on the assumption that a successful method should maintain the image content and be more reliant on the input image~\cite{xie2020self,zhang2019harmonic}. To enforce the generator learn the input contents we build a graph on the semantic segments extracted from the image. This encourages generator to maintain the objects as well as their relationship during translation. For generating more diverse and higher quality images, we introduce style reference image and maximize the similarity between the output image and reference image. By maximizing the mutual information between the matched objects, we ensure the generator inherits the significant representations that are crucial for image creation. This enforces the generator to produce more diverse and realistic images. Finally, our model allows user control over the style of input image using style images. 

\section{\name}
\par\textbf{Problem Definition:}
 \textcolor{black}{Given the input domain $\mathcal{X}$, the output domain $\mathcal{Y}$, a reference style image~$\mathbf{S}$, trainable parameters $\theta$, and a set of unpaired examples $\mathbf{X} \in \mathbb{R}^{H\times W\times C}$ and $\mathbf{Y} \in \mathbb{R}^{H\times W\times C}$, the goal is to learn a mapping function~$f(\mathbf{X}, \mathbf{Y}, \mathbf{S};\theta):\mathbb{R}^{H\times W\times C} \xrightarrow[]{}\mathbb{R}^{H\times W\times C}$ from $\mathcal{X}$ to $\mathcal{Y}$.
This mapping function is an extension to a typical image-to-image translation as defined in~\cite{zhang2019harmonic,zhu2017unpaired,abbasnejad2020gold,liu2016coupled,abbasnejad2017using} where $f(\textbf{X}, \textbf{Y};  \theta)$ maps~$\textbf{X}$ to a new domain $\textbf{Y}$ using a Gaussian noise distribution.
However, synthesizing a scene under new conditions cannot be easily controlled by using only a Gaussian noise vector, since more realistic and diverse image generation is required in the output domain. In our work, we introduce style image $\mathbf{S}$ for image manipulation, with the goal to synthesise images in the target domain using a reference style image, control the style and content of $\mathbf{Y} \in \mathcal{Y}$.
For more realistic and diverse image generation we consider $\mathbf{S}$ having mutual properties as the input image (for example it is drawn from the same dataset as $\{\mathbf{X, Y}\}$); however, we can ease this constraint and assume $\mathbf{S}$ is coming from any arbitrary distribution.}
 
 \par \textbf{Overview of Approach:~}\textcolor{black}{ As explained earlier, a successful image translator should maintain the image structure in $\mathbf{X} \in \mathcal{X}$ and $\mathbf{Y} \in \mathcal{Y}$. To encourage the generator to maintain the structural relationship between the objects, we build a graph on the semantic segments extracted from the input and output images. For generating more diverse and higher quality images, we introduce style reference image and maximize the mutual information between the matched objects in the output and the style reference image. This encourages the generator to inherit the significant features for stylizing images in the new domain. We generate style-vector from $\mathbf{S}$ to reflect the feature vector in the output images. There are two benefits in this setting: (i) during training the generator learns to output diverse images; (ii) it enables users to control the style of images in the target space. For example, not only can a user translate images between domains e.g. summer2winter, but she can also control the style of the output image (e.g., less snow on the mountain and more on the ground).}
 
 
 \subsection{Network architecture}
 This section explains details regarding the modules that are used in this work.
 \\
 \textbf{Generator:}
 Since the generator needs to learn style reference image characteristics as well as synthesising images using random noise, we feed the input image $\mathbf{x} \in \mathbf{X}$, style-vector $\mathbf{s} \in \mathbb{R}$ and the latent code$~\mathbf{z}\in \mathbb{R}$ to the generator. 
The generator consists of four down-sampling blocks, four intermediate blocks, and four up-sampling blocks, which use pre-activation residual units~\cite{he2016identity}. We use the instance normalization (IN)~\cite{ulyanov2016instance} while down-sampling and utilize the adaptive instance normalization (AdaIN)~\cite{huang2017arbitrary} for up-sampling blocks. The style-vector and latent code are concatenated and injected into all AdaIN layers.
\\
\textbf{Style encoder:} The encoder maps the reference style image to the style-vector $\mathbf{s} \in \mathbb{R}$. The encoder has a CNN layer and six pre-activation residual units~\cite{he2016identity} followed by two branches of fully connected layer size $64 \times 2$. We set the number of branches as two (binary translation). 
\\
\textbf{Latent mapping network:} The input of the mapping network is Gaussian noise vector and a domain. The output of the mapping encoder is a vector that feeds into the generator. The mapping network has eight fully-connected layers that takes the input vector and outputs a vector of size $64\times 2$.
\\
\textbf{Discriminator:} The discriminator has a CNN layer and six pre-activation residual units~\cite{he2016identity} and a fully-connected layer that is applied to the last residual block. The output of discriminator is a real/fake classifier for each branch.

\subsection{Learning spatial dependencies}
As mentioned earlier, a successful image-to-image translator should maintain the image structure while synthesizing the input image in the new domain~\cite{zhang2019harmonic,bashkirova2019adversarial}. Therefore, learning the spatial relationship between the objects in the input image is vital for an accurate and realistic image generation. We learn the image structure using graph convolutional networks (GCN) which have been widely used in different computer vision problems~\cite{garcia2017few,nguyen2018graph}. 
See~\cite{zhou2018graph} for a recent survey on models and applications of graph convolutional networks.

\subsubsection{Graph Convolutional Networks}
\label{sec:gcn}
Typically graph convolutional networks (GCN), $\mathcal{G}(V, \mathbf{A})$ can be defined as a set of vertex nodes $V = \{v_{1}, ..., v_{n}\}$ and $\mathbf{A} \in \mathbb{R}^{n\times n}$, a symmetric (typically sparse) adjacency matrix. 
Each node $v_{i}$ in the graph has a corresponding $d$-dimensional feature vector obtained from a linear operation on the $k-$th input $\mathbf{Q}^k =[\mathbf{q}^k_1, ..., \mathbf{q}^k_{n}] \in \mathbb{R}^{n\times d}$ where $\mathbf{Q}^0$ is the input and $k\in\{0,\ldots,L\}$ for each layer $L$. 
Given the input to a GCN layer $\mathbf{Q}^{k} \in \mathbb{R}^{n\times d}$, it produces $\mathbf{Q}^{k+1} \in \mathbb{R}^{n\times d_{k+1}}$ as the $k+1$-th output of the GCN$(.)$ layer:
\begin{align}
\label{eq:gcn_output}
          \mathbf{Q}^{k+1} = \rho( \mathbf{A} \mathbf{Q}^{k} \boldsymbol\Theta ),
\end{align}
where $\rho$ is a non-linearity function (Leaky ReLU~\cite{xu2015empirical} in our case) and $\boldsymbol\Theta  \in \mathbb{R}^{d_{k} \times d_{k+1}}$ is a set of GCN parameters. For simplicity we consider adjacency operator 
learned as:
\begin{align}
\label{eq:adjacency_op}
          {A}_{i, j}^{k} = \phi_{}(|\mathbf{q}_{i}^{k} - \mathbf{q}_{j}^{k}|), 
\end{align}
where $\phi$ is a symmetric function. Here we consider a neural network that has $5$ layers of convolutional layers and a fully connected layer which works on the absolute difference between two feature vectors. ${A}_{i, j}^{k} = 1$ if $i = j $ implies each vertex in the graph is self-connected. The final graph would be obtained by stacking Eq.~\ref{eq:gcn_output} for $L$ times:
\begin{align}
\label{eq:graph}
          \mathcal{G}(\mathbf{Q};\boldsymbol\Theta) = \sigma(\text{GCN}_L(... \text{GCN}_1(\mathbf{Q}^{0}))),
\end{align}
where $\sigma$ is a softmax function. 

\begin{figure*}[t!]
\vspace{-15pt}
\begin{center}
   \includegraphics[width=0.99\linewidth]{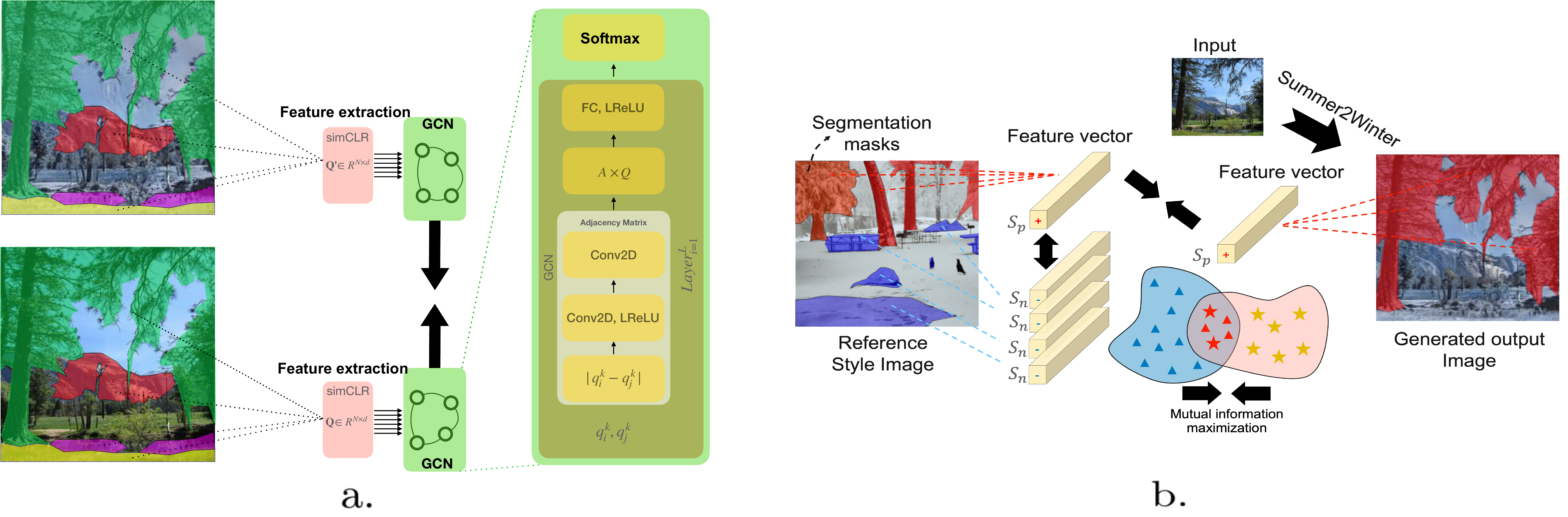}
\end{center}
\vspace{-15pt}
   \caption{\textbf{a.} The graph convolutional network which learns the structures among the objects. We build two GCN on the input image and generated image. We use simCLR~\cite{chen2020simple} to extract features from the segmented images., \textbf{b.} Our mutual information maximization, tries to learn the feature contents from the ”Reference Style Image” in order to generate more realistic and diverse images. We segment the objects from both reference image and the output. We put weight on the objects that are similar and penalize the objects that are different.}
\label{fig:graph_conv}
\end{figure*}[t!]
\subsubsection{Graph convolution for spatial dependencies}
\label{sec:gcn_spa}
As introduced in Section~\ref{sec:gcn}, a GCN can model the relations among different objects and learn powerful representations for object localization. Therefore, we use a GCN module for learning the spatial relations between the objects in the input and generated images.

We assume $\mathbf{Q}^{0}$ as the input signal to a GCN$_1$ module. We select $n$ objects from the input image and pass them to a constractive learning module. \textcolor{black}{We use a segmentation model to extract $n$ objects from each image. Objects are selected as the most prominent objects per dataset}. We use simCLR~\cite{chen2020simple} which learns representations that are invariant under a set of augmentations through a contrastive loss. The intuition behind using a contractive learning framework is that, the output of an image-to-image translator must be a variation of the input image, therefore simCLR should consider the extracted segments from input and output as different views of same objects and should generate same $d-$dimensional features. \textcolor{black}{If an object doesn't exist in the image a vector of zeros is put as the extracted features.} We build two GCN on the input and output images (see Figure~\ref{fig:graph_conv}a. for an example) and minimize the following loss for learning the spatial dependencies between the objects:

\begin{align}
\label{eq:adv_spatialloss}
          \mathcal{L}_{spatio} =  \| &\mathcal{G}(\mathbf{Q}; \boldsymbol\Theta) -  \mathcal{G}(\mathbf{Q}'; \boldsymbol\Theta')\|_{2}^{2},
\end{align}
where $\mathbf{Q}$ and $\mathbf{Q}'$ are the features extracted from $n$ segmented objects from input and output images respectively. 

\subsection{Mutual information maximization}
\label{sec:mim}
As explained earlier, we intend to introduce a robust, realistic and diverse image generator. GCN can maintain the image structure between the input and output images, however it doesn't capture the necessary information (e.g. texture, color and etc.) for generating diverse images. In addition, we would like~\name~learns the style of reference image for image manipulation in the target domain. As introduced in~\cite{hjelm2018learning} mutual information maximization between the reference image and generated output can improve the reconstruction quality. With mutual information maximization, the generator’s prediction distribution will be balanced and better reflect the style of reference image in the output space. Inspired by~\cite{li2020rethinking} we use mutual information maximization for our image-to-image translation. Followed by~\cite{li2020rethinking} the mutual information maximization can be broken in two parts: $I({\mathbf{S}}; \mathbf{Y})  = H(\mathbf{S}) - H(\mathbf{Y}|{\mathbf{S}})$, maximizing the info-entropy $H(\mathbf{Y})$ and minimizing the conditional entropy $H(\mathbf{Y}|{\mathbf{S}})$, where $\mathbf{Y}\in \mathcal{Y}$ is the generated image. 
This objective will enforce the generator to generate more diverse images in the target domain~\cite{cui2020towards}. 
 As is shown in~\cite{oord2018representation} Noise-Contrastive Estimation (NCE) is a lower bound for maximizing mutual information. Therefore for maximizng the mutual information between the style image and the output we use the following loss function:
\begin{align}
\label{eq:infomax}
          \mathcal{L}_{info} (\mathbf{Y}) = 
          \mathbb{E}_{\mathbf{x}}  \left[-\log\left(\frac{\sum_{j=1}^{S_{p}}e^{\tilde{\mathbf{s}}_{j}\frac{\tilde{\mathbf{y}}^{+}}{\eta}}}{\sum_{j=1}^{S_{p}}e^{\tilde{\mathbf{s}}_{j}.\frac{\tilde{\mathbf{y}}^{+}}{\eta}} + \sum_{j=1}^{S_{n}}e^{\tilde{\mathbf{s}}_{j}.\frac{\tilde{\mathbf{y}}^{-}}{\eta}}}\right)\right],
\end{align}
where $\eta$ is a constant, $S_{p}$ and $S_{n}$ correspond to the set of positive and negative samples extracted from the encoded reference style image $\tilde{\mathbf{s}}$ and encoded generated image $\tilde{\mathbf{y}}$. \textcolor{black}{Positive samples are defined as those objects that are same in the style image and output image, and negative ones are those that are different.} In other words, we put weights on the objects that are mutual in the style reference image and the output and penalized the objects that are different. Figure~\ref{fig:graph_conv}b. visualizes the concept of our proposed mutual information maximization setting.
\subsection{Final objective}
\label{sec:final_obj}
In addition to the objectives we presented in Section~\ref{sec:gcn_spa} and Section~\ref{sec:mim}, we use the \textit{Adversarial loss, Cycle consistency} and \textit{Style modelling loss}.
\\
\textbf{Adversarial loss:}
We use the adversarial loss~\cite{goodfellow2014generative} for training the generator and discriminator simultaneously:
\begin{align}
\label{eq:adv_loss}
          \mathcal{L}_{adv} (G, D) = \mathbb{E}_{\mathbf{x}, y} [ \log D (\mathbf{x})] &+ \\ \nonumber 
           \mathbb{E}_{\mathbf{x}, {y}, \mathbf{z}} &[ \log (1 &- D_{{y}} (G(\mathbf{x}, \hat{\mathbf{s}})))],
\end{align}
where $G$ tries to minimize this objective against an adversarial $D$ that tries to maximize it. In Eq.~\ref{eq:adv_loss}, $\hat{\mathbf{s}}$ is the output of the style encoder given 
the style image, $\mathbf{S}$, and target domain $y$. During training, the generator learns to predict images that are indistinguishable from the real images of domain $y$ given the style image $\mathbf{S}$.
\\
\textbf{Cycle consistency loss:} 
\textcolor{black}{In image-to-image translation to learn the mapping between the unpaired input and output images a cycle-consistency loss is required~\cite{zhu2017unpaired}:}
\begin{align}
\label{eq:cycle_loss}
          \mathcal{L}_{cycle}  =  \| \mathbf{x} - G(G(\mathbf{x}, \hat{\mathbf{s}}), \tilde{\mathbf{s}})\|_{1},
\end{align}
where $\tilde{\mathbf{s}} = E(\mathbf{x})$ is the output of the encoder given the input image $\textbf{x}$. 
\\
\textbf{Style modelling:} As mentioned in Section~\ref{sec:mim}, the generator is encouraged to generate realistic and diverse images. However, style reference image and the output may have a very little mutual information. This may result very large values in Eq.~\ref{eq:infomax} and promotes mode collapse for the generator. To reduce this behaviour we introduce style modelling loss as follows:
\begin{align}
\label{eq:style_model_loss}
          \mathcal{L}_{style}  =  \| \hat{\mathbf{s}} -  E_{y}(G(\mathbf{x}, \hat{\mathbf{s}}))\|_{1},
\end{align}
\\
\textbf{Final objective:} Finally we optimize the weight summation of the losses in Section~\ref{sec:gcn}, Section~\ref{sec:mim} and Section~\ref{sec:final_obj}. The entire loss function is defined as follows:  
\begin{align}
\label{eq:total_loss}
\min_{G, E, M} &\max_{D} \lambda_{adv}\mathcal{L}_{adv} + \lambda_{spatio}\mathcal{L}_{spatio} + \\ \nonumber
&\lambda_{info}\mathcal{L}_{info} +\lambda_{cycle}\mathcal{L}_{cycle} + \lambda_{style}\mathcal{L}_{style}
\end{align}
where $\lambda$ are a set of hypo-parameters that are tuned manually during training.

\begin{figure*}[t!]
\vspace{-10pt}
\begin{center}
\begin{tabular}{C{25pt}C{425pt}}
    \multicolumn{1}{c}{} &{\multirow{28}{*}{\includegraphics[width=14.3cm,height=12.3cm]{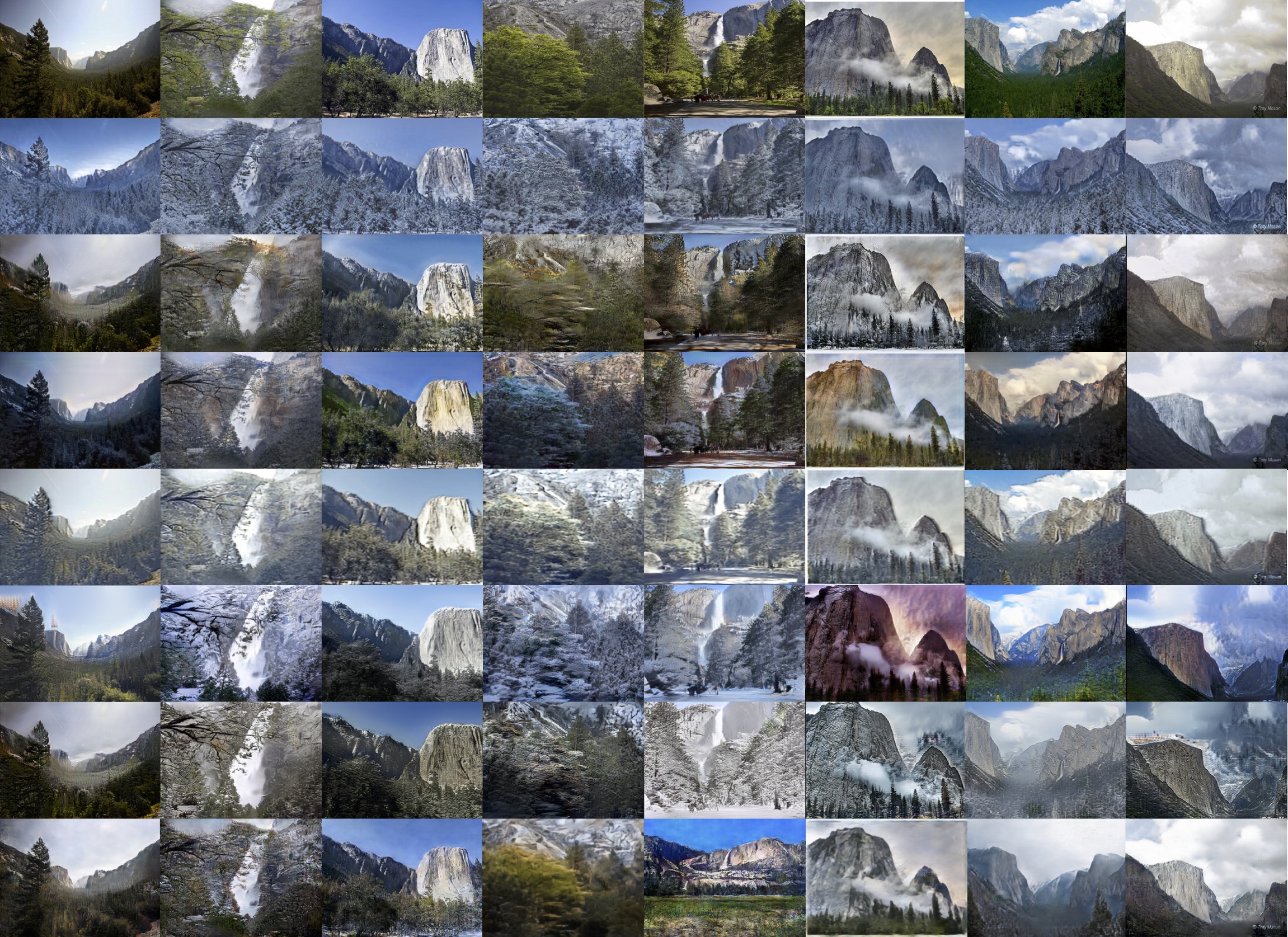}}}\\
    \multicolumn{1}{c|}{\tiny{\textbf{Input}}}&{}\\
    \multicolumn{1}{c|}{ }&{}\\
    \multicolumn{1}{c|}{ }&{}\\
    \multicolumn{1}{c|}{ }&{}\\
    \multicolumn{1}{c|}{\tiny{SCONE-GAN}}&{}\\
    \multicolumn{1}{c|}{ }&{}\\
    \multicolumn{1}{c|}{ }&{}\\
    \multicolumn{1}{c|}{ }&{}\\
   \multicolumn{1}{c|}{\tiny{CUT~\cite{park2020contrastive}}}&{}\\
    \multicolumn{1}{c|}{ }&{}\\
    \multicolumn{1}{c|}{ }&{}\\
    \multicolumn{1}{c|}{\tiny{Cycle-GAN~\cite{zhu2017unpaired}}}&{}\\
    \multicolumn{1}{c|}{ }&{}\\
    \multicolumn{1}{c|}{ }&{}\\
    \multicolumn{1}{c|}{ }&{}\\
    \multicolumn{1}{c|}{\tiny{MUNIT~\cite{huang2018multimodal}}}&{}\\
    \multicolumn{1}{c|}{ }&{}\\
    \multicolumn{1}{c|}{ }&{}\\
    \multicolumn{1}{c|}{ }&{}\\
    \multicolumn{1}{c|}{\tiny{DRIT++~\cite{lee2020drit++}}}&{}\\
    \multicolumn{1}{c|}{ }&{}\\
    \multicolumn{1}{c|}{ }&{}\\
    \multicolumn{1}{c|}{\tiny{MSGAN~\cite{mao2019mode}}}&{}\\
    \multicolumn{1}{c|}{ }&{}\\
    \multicolumn{1}{c|}{ }&{}\\
    \multicolumn{1}{c|}{ }&{}\\
    \multicolumn{1}{c|}{\tiny{CyCADA~\cite{hoffman2018cycada}}}&{}\\
    \multicolumn{1}{c|}{ }&{}\\
\end{tabular}
\end{center}
\vspace{-15pt}
\caption{Examples of generated images using our proposed framework. The top row shows the original images in summer domain and the lower images shows the generated winter image using different methods.}
\label{fig:results_compare}
\end{figure*}

\section{Optimization and Inference}
\label{sec:optim}
For network training similar to~\cite{goodfellow2014generative}, instead of minimizing $\log(1 - D_{y}(G(\mathbf{x}, \hat{\mathbf{s}}))$ for training $G$, we maximize $\log D(\mathbf{x}, G(\mathbf{x},\hat{\mathbf{s}}))$. We also use $R_{1}$ regularization~\cite{mescheder2018training} with $\gamma = 1$. Furthermore, we set $\lambda_{adv} = 1$, $\lambda_{spatio} = 1$, $\lambda_{info} = 2$, $\lambda_{cycle} = 2$ and $\lambda_{style} = 2$. We use the Adam optimizer with $\beta_{1} = 0$ and $\beta_{2} = 0.99$. The learning rates for generator, discriminator, and the encoder, are set to $10^{-4}$, and for the mapping network is set to $10^{-6}$. We initialize all weights of the convolutional, fully-connected, and affine transform layers using $\mathcal{N} (0, 1)$. The biases and noise scaling factors are initialized to zero, except biases associated with the scaling vectors of AdaIN that are set to one. For our encoder, we use leaky ReLU with $\alpha = 0.2$~\cite{maas2013rectifier} and equalized learning rate~\cite{karras2017progressive} for all layers. We do not use batch normalization, spectral normalization, attention mechanisms, dropout, or pixelwise feature vector normalization~\cite{karras2019style}.

\section{Experiments} 
This section provides details regarding the datasets we used for evaluation, the evaluation metrics and the results.

\textcolor{black}{To explore the performance of SCONE-GAN, we evaluate our method on a variety of unpaired datasets that contain several objects with different shapes}:
\\
\textbf{Yosemite dataset:} This dataset contains $854$ winter images and $1273$ summer images of Yosemite national park~\cite{zhu2017unpaired}.
\\
\textbf{Monet2photo dataset:} This dataset has $1072$ images of painting and $6287$ images of landscape~\cite{zhu2017unpaired}. 
\\
\textbf{Nordlandsbanen dataset:}~\cite{sunderhauf2013we} This dataset consists videos from a train journey on the same railway track once in every season (spring, summer, autumn, winter). Each video features seasonal effects like snow, color changing foliage and different weather and lighting conditions. We select $5500$ unpaired frames from the summer and winter videos for training and $280$ paired images for testing.
\\
\textbf{Cityscape dataset:} The dataset contains $2975$ training and $500$ testing images from the Cityscapes dataset~\cite{cordts2016cityscapes}. 
{
\begin{table*}[t!]
\vspace{-10pt}
\begin{center}
\centering
\small
\begin{tabular}{C{48pt}|C{22pt}|C{24pt}|C{22pt}|C{22pt}|C{22pt}|C{24pt}|C{22pt}|C{22pt}|C{22pt}|C{24pt}|C{22pt}|C{22pt}}
\multicolumn{1}{c|}{ } & \multicolumn{4}{c|}{\textbf{\small Yosemite}~\cite{zhu2017unpaired}} &  \multicolumn{4}{c|}{\textbf{\small Cityscape}~\cite{cordts2016cityscapes}} & \multicolumn{4}{c}{\textbf{\small Monet2photo}~\cite{zhu2017unpaired}} \\
\cline{1-13}
\scriptsize {Method} & \scriptsize{FID} & \scriptsize{LPIPS}& \scriptsize{NDB}& \scriptsize{JSD}  & \scriptsize{FID} & \scriptsize{LPIPS}& \scriptsize{NDB}& \scriptsize{JSD}& \scriptsize{FID}& \scriptsize{LPIPS}& \scriptsize{NDB}& \scriptsize{JSD} \\
\hline
\hline
\scriptsize CUT~\cite{park2020contrastive}        & \scriptsize64.17 & \scriptsize0.31 & \scriptsize31.74 & \scriptsize0.049& \scriptsize57.14 & \scriptsize0.14& \scriptsize27.05& \scriptsize0.029& \scriptsize66.51& \scriptsize0.28& \scriptsize28.79& \scriptsize0.085\\
\scriptsize DRIT++~\cite{lee2020drit++}        & \scriptsize61.12 & \scriptsize0.24 & \scriptsize28.03 & \scriptsize0.056 & 
\scriptsize72.47 & \scriptsize0.17 & \scriptsize22.62& \scriptsize0.049&
\scriptsize71.65& \Blue{\scriptsize0.53}& \Blue{\scriptsize21.14}& \scriptsize0.087\\
\scriptsize MUNIT~\cite{huang2018multimodal}        & \scriptsize66.64 & \scriptsize0.21 & \scriptsize33.61 & \scriptsize0.062& 
\scriptsize84.53& \scriptsize0.22 & \scriptsize34.54& \scriptsize0.065&
\scriptsize72.42& \scriptsize0.51& \scriptsize23.47& \scriptsize0.091\\
\scriptsize CycleGAN~\cite{zhu2017unpaired}        & \scriptsize71.20 & \scriptsize0.13 & \scriptsize45.73 & \scriptsize0.057&
\scriptsize76.30& \scriptsize0.13 & \scriptsize29.67& \scriptsize0.058& \scriptsize77.85 & \scriptsize0.35& \scriptsize26.78& \scriptsize0.098\\
\scriptsize MSGAN~\cite{mao2019mode}        & \scriptsize52.65 & \scriptsize0.13 & \scriptsize25.30 & \scriptsize0.042& 
\scriptsize79.15 & \Blue{\scriptsize0.26} & \scriptsize23.61 & \scriptsize0.034 & \scriptsize71.73 & \scriptsize0.41 & \scriptsize24.65 & \scriptsize0.079 \\
\scriptsize CyCADA~\cite{hoffman2018cycada}        & \scriptsize54.69 & \scriptsize0.33 & \scriptsize25.18 & \scriptsize0.034& 
\scriptsize69.31 & \scriptsize0.20 & \scriptsize21.15 & \scriptsize0.028 & \scriptsize68.41 & \scriptsize0.42 & \scriptsize27.46 & \scriptsize0.075 \\
\hline
\scriptsize SCONE-GAN        & \Blue{\scriptsize51.70} & \Blue{\scriptsize0.40} & \Blue{\scriptsize22.51} & \Blue{\scriptsize0.031}& 
\Blue{\scriptsize56.37} & {\scriptsize0.19} & \Blue{\scriptsize20.54}& \Blue{\scriptsize0.021}& 
\Blue{\scriptsize63.24}& \scriptsize0.49& \scriptsize22.76& \Blue{\scriptsize0.074}\\

\hline
\end{tabular}
\end{center}
\vspace{-15pt}
\caption{Quantitative results for the Yosemite, Cityscape and Monet2photo datasets. We report FID (lower is better~$\downarrow$), LPIPS (higher is better~$\uparrow$)~NDP (lower is better~$\downarrow$), and JSD (lower is better~$\downarrow$).}
\label{tbl:results_tbl}
\end{table*}}
\subsection{Evaluation Metrics}
We use the following evaluation metrics:
\\
\textbf{FID:} To evaluate the quality of the generated images, we use FID~\cite{heusel2017gans}. FID measures the distance between the generated distribution and the real one through the extracted features by Inception Network~\cite{szegedy2015going}. Lower FID values indicate better quality of the generated images.
\\
\textbf{LPIPS:} We use LPIPS ~\cite{zhang2018unreasonable} to evaluate the diversity of the generated images. LIPIS measures the average feature distances between generated samples. Higher LPIPS score indicates better diversity among the generated images.
\\
\textbf{NDB} and \textbf{JSD:} Measure the similarity between the distribution between real images and generated samples~\cite{richardson2018gans} and also evaluate the extent of mode missing of generative models. Following~\cite{richardson2018gans}, the training samples are first clustered using K-means into different bins. Then each synthesized sample is assigned to the bin of its nearest neighbor. Then the bin-proportions of the training samples and the synthesized samples are calculated to evaluate the difference between the generated distribution and the training distribution. 
 Lower NDB and JSD values  mean the generated data distribution approaches the real data distribution better.

\subsection{Setup}
\label{sec:setup}
\textcolor{black}{
During training we use data augmentation. We flip images horizontally with a probability of $0.5$. We resize all the images to $256 \times 256$, the batch size is set to $8$ and all models are trained for $100$K iterations. We use~\cite{lambert2020mseg} method for segmenting the images.} 


\subsection{Results}

\begin{figure}[t]
\vspace{-10pt}
     \centering
     \begin{subfigure}[b]{0.49\textwidth}
         \centering
         \includegraphics[width=8.0cm,height=3.10cm]{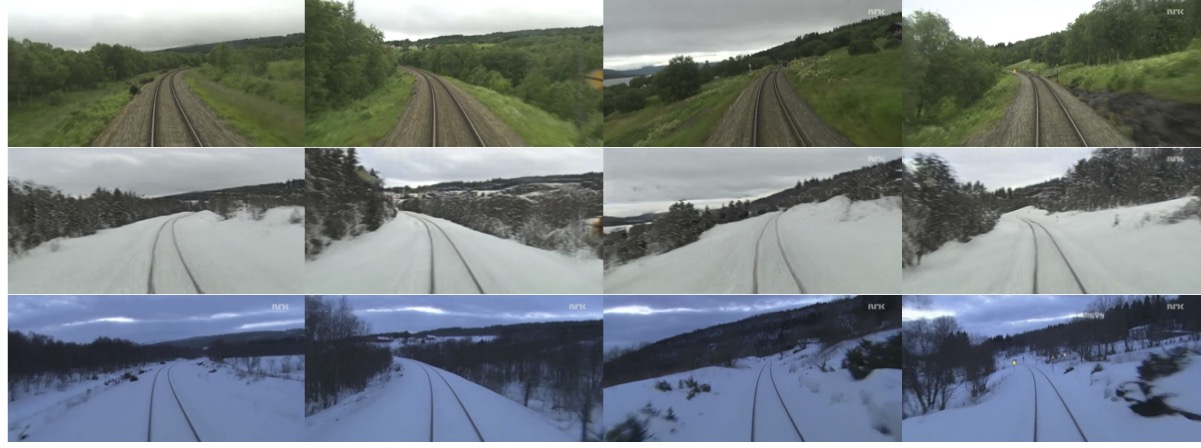}
         \caption{Nordlandsbanen dataset}
         \label{fig:results_train}
     \end{subfigure}
     \hfill
     \begin{subfigure}[b]{0.49\textwidth}
         \centering
         \includegraphics[width=8.0cm,height=3.10cm]{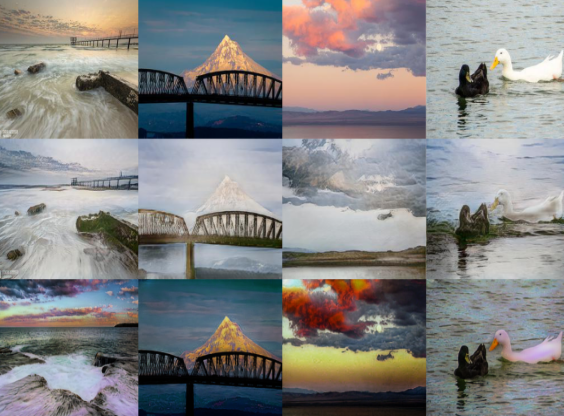}
         \caption{Monet painting dataset}
         \label{fig:mono_results}
     \end{subfigure}
     \hfill
     \vspace{-10pt}
     \caption{\textbf{a.} Examples of generated images on the Nordlandsbanen dataset. The top row shows the input images. The middle row shows the generated images in the winter domain and the third row shows the ground truth (recorded by a camera in winter). \textbf{b.} Transferring images into Monet painting photos. First row is the original image, second row and third row show the transformed images using SCONE-GAN and CUT~\cite{park2020contrastive} respectively.}
\end{figure}

\textbf{Qualitatively:}
First we qualitatively compare our method with six baselines MUNIT~\cite{huang2018multimodal}, CycleGAN~\cite{zhu2017unpaired}, DRIT++~\cite{lee2020drit++},  CUT~\cite{park2020contrastive}, MSGAN~\cite{mao2019mode} and CyCADA~\cite{hoffman2018cycada}. \textcolor{black}{Since previous approaches mostly synthesised images in the output domain given a noise variable, in this experiment we do not use any reference image.} Figure~\ref{fig:results_compare} compares SCONE-GAN with the introduced baselines for summer2winter evaluation. Our method encourages the generator to preserve the image structure while transforming images between domains. Therefore, during translation, the artifacts are minimum in the output images. In addition, we enforce the model to maximize the mutual information between the output and reference images for enhancing the diversity of generated images in the output space. As is shown in Figure~\ref{fig:results_compare}, the generated images by SCONE-GAN are almost more meaningful and have less artifacts than other methods and better representing an image in winter domain (the trees, mountain and ground are better covered by snow).
We also test SCONE-GAN on Nordlandsbanen dataset~\cite{sunderhauf2013we}. The results for this experiment are illustrated in Figure~\ref{fig:results_train}. For this experiment we compare our results with the ground truth images which have been captured in winter. As is visualized in Figure~\ref{fig:results_train} the generated results are realistic, and statistically similar to the real images. We have also compared our model on Monet2photo dataset~\cite{zhu2017unpaired}. The generated results for this experiment is shown in Figure~\ref{fig:mono_results}. For this evaluation we compare our model with "CUT"~\cite{park2020contrastive} which have obtained state-of-the-art results on Monet2photo dastset~\cite{park2020contrastive}. \textcolor{black}{As can be seen from the figure, SCONE-GAN preserves the image content and better translates the input image in the target domain.}

\textbf{Quantitative:}
As the quantitative results exhibited in Table~\ref{tbl:results_tbl}, SCONE-GAN is outperforming the selected baseline in almost all metrics. 
We obtain the lowest FID~\cite{heusel2017gans} against the baseline, suggesting that the generated images are more realistic and have better image quality. We have also achieved improvements on LPIPS~\cite{zhang2018unreasonable}, NDB and JSD ~\cite{richardson2018gans} values in comparison with the baseline confirming that our method can generate more diverse images. We observed that for Cityscape~\cite{cordts2016cityscapes} and Monet2photo~\cite{zhu2017unpaired} datasets the segmentation module~\cite{lambert2020mseg} sometimes fail to extract the accurate objects. In addition, simCLR~\cite{chen2020simple} did not generate robust features on some classes because it was not trained on such images. 


\subsection{Amazon Mechanical Turk Experiment}
We conducted a user study through Amazon Mechanical Turk where users were asked to measure the perceptual realism of the generated images. We show 180 images from Yosemite test dataset~\cite{zhu2017unpaired}, 90 real and 90 fake, to 50 participants and asked to distinguish real from fake. We report average classification Accuracy$_{\text{time}}$ score, the minimum time in second that participants need to see an image and classify it as a real or fake image, in Table~\ref{tbl:amt}. In this experiment, we compare our method with CyCADA~\cite{hoffman2018cycada} and CUT~\cite{park2020contrastive}. In the gray columns we report the average classification accuracy of all participants. To have a better comparison and ensure that the participants have a better judgment, we choose the participants who have gained average classification accuracy of $80\%$ or above on the real images. This experiment is shown in the cyan color column. As can be seen from this experiment, SCONE-GAN has obtained better results on human perceptual evaluation. In this table, higher classification accuracy on real images shows the participants were more successful on classifying real images and lower classification accuracy on fake images shows the method successfully fooled the participants.
{
\begin{table}[t!]
\vspace{-5pt}
\begin{center}
\centering
\small
\begin{tabular}{C{30pt}|C{12pt}|C{12pt}|C{12pt}|C{12pt}|C{12pt}|C{12pt}|C{12pt}|C{12pt}}
\multicolumn{1}{c|}{ } & 
\multicolumn{2}{>{\columncolor{LightCyan}}c|}{\tiny{Accuracy$_{20} \%$}} &  \multicolumn{2}{>{\columncolor{LightCyan}}c|}{\tiny{Accuracy$_{\infty}\%$}} & \multicolumn{2}{>{\columncolor{gray}}c|}{\tiny{Accuracy$_{20} \%$}} &
\multicolumn{2}{>{\columncolor{gray}}c}{\tiny{Accuracy$_{\infty} \%$}} \\
\cline{1-9}
 \tiny {Method} &  \tiny{Real }& \tiny{Fake }&  \tiny{Real }& \tiny{Fake }& \tiny{Real}& \tiny{Fake }& \tiny{Real}& \tiny{Fake } \\
\hline
\hline
\tiny SCONE-GAN     & \tiny \textbf{\Blue95.82} & \tiny \textbf{\Blue28.26} & 
 \tiny \textbf{\Blue93.54} & \tiny{\textbf{\Blue41.94}} & \tiny{\textbf{\Blue60.13}}& \tiny{\textbf{\Blue47.58}}& \tiny \textbf{\Blue61.04} & \tiny{\textbf{\Blue47.28}}\\
\tiny CUT~\cite{park2020contrastive}      & \tiny \tiny 95.85 & \tiny 40.78 & \tiny 93.25 & \tiny{45.01} &
 \tiny{59.78}& \tiny{51.46}& \tiny 60.64 & \tiny{50.25}\\
\tiny  CyCADA~\cite{hoffman2018cycada}      & \tiny 95.70 & \tiny 30.46 & \tiny 93.68 & \tiny{49.92} &
 \tiny{60.72}& \tiny{51.23}&  \tiny 60.48 & \tiny{50.91}\\
\hline
\end{tabular}
\end{center}
\vspace{-15pt}
\caption{AMT experiment. This table shows the average classification accuracy for SCONE-GAN, CUT~\cite{park2020contrastive} and CyCADA~\cite{hoffman2018cycada}. Higher classification accuracy on real images means the participants were more successful on classifying real images and lower classification accuracy on fake images shows the method were more successful on fooling the evaluators.}
\label{tbl:amt}
\end{table}}
\begin{figure}[t]
\vspace{-5pt}
\begin{center}
   \includegraphics[width=7.2cm, height = 7.9cm]{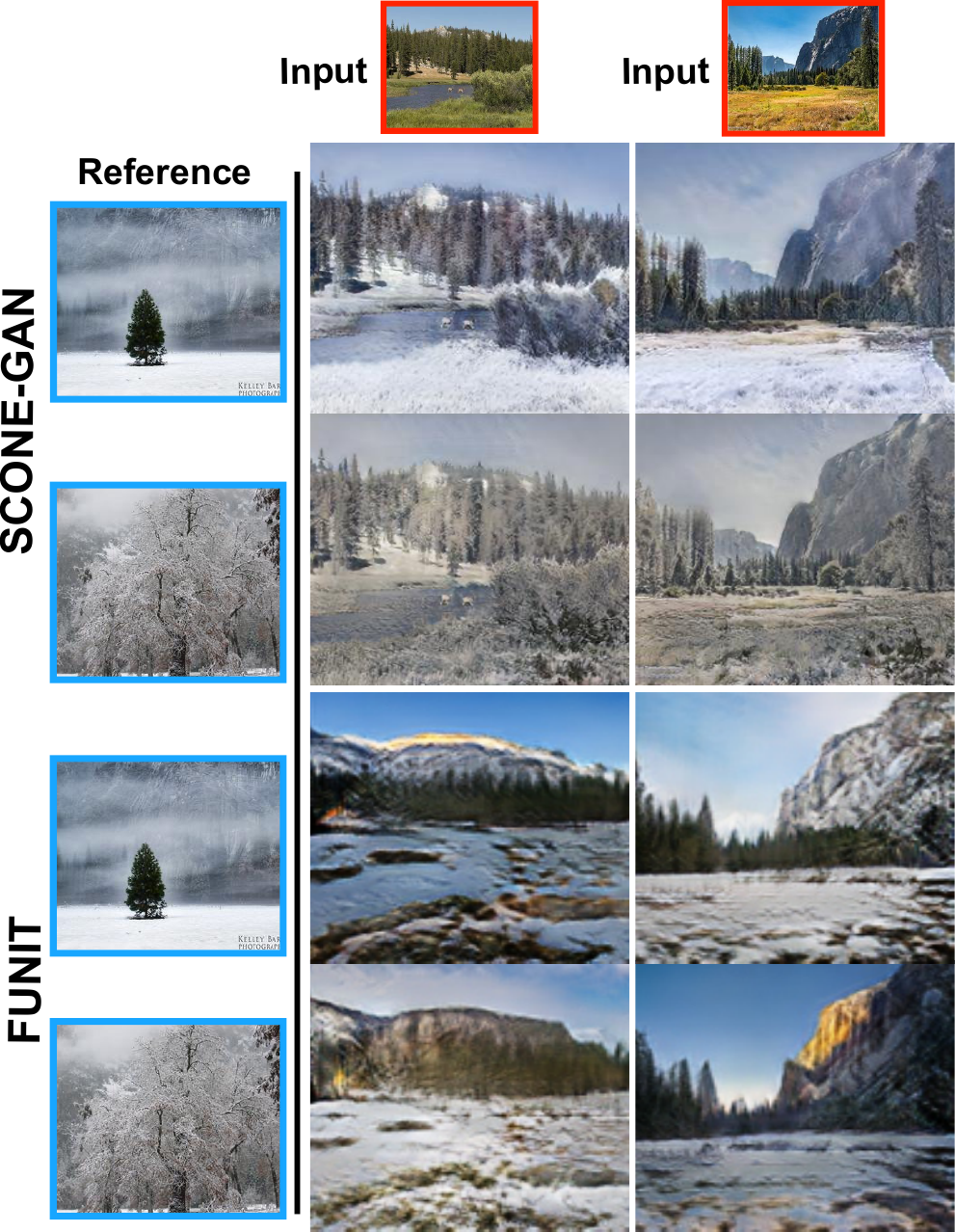}
\end{center}
\vspace{-15pt}
   \caption{Reference-guided image synthesis. Given the input image and a reference style image, SCONE-GAN and FUNIT~\cite{liu2019few} synthesize summer2winter images in the target domain.}
\label{fig:reference_results}
\end{figure}

{
\begin{table*}[t!]
\vspace{-15pt}
\begin{center}
\centering
\small
\begin{tabular}{C{60pt}|C{22pt}|C{22pt}|C{22pt}|C{22pt}|C{22pt}|C{22pt}|C{22pt}|C{22pt}|C{22pt}|C{22pt}|C{22pt}|C{22pt}}
\multicolumn{1}{c|}{ } & \multicolumn{4}{c|}{\textbf{\small Yosemite dataset}~\cite{zhu2017unpaired}} &  \multicolumn{4}{c|}{\textbf{\small Cityscape dataset}~\cite{cordts2016cityscapes}} & \multicolumn{4}{c}{\textbf{\small Monet2photo dataset}~\cite{zhu2017unpaired}} \\
\cline{1-13}
 \scriptsize {Method} & \scriptsize{FID} & \scriptsize{LPIPS}& \scriptsize{NDB}& \scriptsize{JSD}  & \scriptsize{FID} & \scriptsize{LPIPS}& \scriptsize{NDB}& \scriptsize{JSD}& \scriptsize{FID}& \scriptsize{LPIPS}& \scriptsize{NDB}& \scriptsize{JSD} \\
\hline
\hline
\scriptsize \textcolor{red}{$\lambda_{2}=0$}, $\lambda_{3, 5}=2$       & \scriptsize63.96 & \scriptsize0.25 & \scriptsize39.74 & \scriptsize0.049& 
\scriptsize68.46 & \scriptsize{0.21} & \scriptsize{24.13}& \scriptsize{0.038}& \scriptsize{69.17}& \scriptsize0.39 & \scriptsize25.14 & \scriptsize{0.084}\\
\scriptsize  $\lambda_{2}=1$, \textcolor{red}{$\lambda_{3, 5}=0$}       & \scriptsize68.54 & \scriptsize0.22 & \scriptsize42.79 & \scriptsize0.051& \scriptsize70.12 & \scriptsize{0.23} & \scriptsize{27.92}& \scriptsize{0.052}& \scriptsize{71.43}& \scriptsize0.36 & \scriptsize27.44 & \scriptsize{0.087}\\
\scriptsize  \textcolor{red}{{$\lambda_{2, 3, 5}=0$}}       &\scriptsize 70.81 &\scriptsize 0.14 & \scriptsize44.69 & \scriptsize0.055& \scriptsize77.46 & \scriptsize{0.23} & \scriptsize{28.45}& \scriptsize{0.061}& \scriptsize{75.24}& \scriptsize0.34 & \scriptsize28.16 &\scriptsize {0.089}\\
\hline
\end{tabular}
\end{center}
\vspace{-10pt}
\caption{Ablation experiments on components of the SCONE-GAN. $\lambda_{2,3,5}$ refers to $\lambda_{spatio, info, style}$. By setting $\lambda_{2} = 0$ we are cancelling the GCN method and by putting $\lambda_{3} = 0$ and $\lambda_{5} = 0$ we are removing mutual information maximization block from SCONE-GAN.}
\label{tbl:ablation_tbl}
\end{table*}}
\subsection{Reference-guided Image Synthesis}
We can also use SCONE-GAN for synthesising images in a new domain conditioned on a style image. We first encode style image, $\mathbf{S}$, to a reference style-vector and then feed the features into the generator for producing images in the new domain. By enforcing the generator to maximize the mutual information between the generated and reference images, our generator learns to produce more realistic and diverse images based on the information given in the reference image. Figure~\ref{fig:reference_results} shows examples from Yosemite dataset~\cite{zhu2017unpaired}. \textcolor{black}{We also compare our results with FUNIT~\cite{liu2019few} method, as they have obtained state-of-the-art results on image translation using style image. As are shown in Figure~\ref{fig:reference_results}, SCONE-GAN produces higher quality images with less artifacts. Also, trees, mountain and ground are better covered by snow and are higher correlated with the reference image.}

\subsection{Ablation study}
We conduct abundant ablation experiments to analyze the components of SCONE-GAN. We follow the same process for training as we explain in Section~\ref{sec:optim} and Section~\ref{sec:setup}. We first explore the effect of GCN~\ref{sec:gcn_spa} on SCONE-GAN by putting $\lambda_{spatio}=0$ in Eq.~\ref{eq:total_loss}. As can be seen from Table~\ref{tbl:ablation_tbl}, the performance of SCONE-GAN drops significantly. Removing GCN block causes the model not to capture the precise dependency among the objects. In addition, we examine the effect of mutual information maximization block by setting $\lambda_{info} = 0, \lambda_{style} = 0$. In this experiment although the objects are preserved during the translation, however the generator doesn't learn to accurately synthesise images in the output space. Finally, we set $\lambda_{spatio} = 0, \lambda_{info} = 0, \lambda_{style} = 0$. In this experiment, we only use Adversarial loss, Eq.~\ref{eq:adv_loss} and Cycle consistency loss, Eq.~\ref{eq:cycle_loss}. Results from Table~\ref{tbl:ablation_tbl} for this experiment confirm that using $\mathcal{L}_{spatio}, \mathcal{L}_{style}$ and $\mathcal{L}_{info}$ can enormously improve the results across different metrics. 

\subsection{Limitations}
SCONE-GAN relies on the contrastive learning and a segmentation frameworks. Therefore, the output quality may be deteriorated as these models failed on the input images. For example as is shown and discussed in Table~\ref{tbl:results_tbl}, for Cityscape~\cite{cordts2016cityscapes} and Monet2photo~\cite{zhu2017unpaired} experiments, segmentation~\cite{lambert2020mseg} and simCLR~\cite{chen2020simple} models failed on some classes. The reason stems from the fact that these models were not initially trained on these tasks (streets or painting photos).

\section{Conclusion}
We introduced SCONE-GAN for generating realistic and diverse scenery images. Our approach enforces the generator to learn the dependency between objects using graph convolutional network to maintain the structure of the images during translation. In order to extend the diversity of generated images we maximize the mutual information between the style image and the output. During training, the generator learns to utilize the necessary information for enhancing the image diversity. Moreover, this presentation enables users to manipulate scenery in different style conditions. Our qualitative and quantitative experiments on four datasets show that the generated images outperforms quality and diversity of the current state-of-the-art.
\section*{Acknowledgement}
We would like to acknowledge our industry partner, Rheinmetall Defence Australia.
None of this research presented in this paper would have been possible without their support in project DIMSIM (Deep Intelligence Machine SIMulator), part of the ACW Program.
{\small
\bibliographystyle{ieee_fullname}
\bibliography{PaperForReview}
}

\end{document}